\documentclass[nohyperref]{article}
\pdfoutput=1

\usepackage{caption}        
\usepackage{subcaption}     
\usepackage{microtype}
\usepackage{graphicx}
\usepackage{booktabs} 
\usepackage[ruled,vlined]{algorithm2e}      

\usepackage{hyperref}

\usepackage[accepted]{icml2022}

\usepackage{amsmath}
\usepackage{amssymb}
\usepackage{mathtools}
\usepackage{amsthm}

\usepackage[capitalize,noabbrev]{cleveref}

\theoremstyle{plain}

\theoremstyle{definition}

\theoremstyle{remark}

\usepackage[textsize=tiny]{todonotes}

\icmltitlerunning{Intelligent Systematic Investment Agent: an ensemble of deep learning and evolutionary strategies}

\begin{document}

\twocolumn[
\icmltitle{Intelligent Systematic Investment Agent: an ensemble of deep learning and evolutionary strategies}



\icmlsetsymbol{equal}{*}

\begin{icmlauthorlist}
\icmlauthor{Prasang Gupta}{equal,comp}
\icmlauthor{Shaz Hoda}{equal,comp}
\icmlauthor{Anand Rao}{comp}
\end{icmlauthorlist}

\icmlaffiliation{comp}{AI and Emerging Technologies, PwC US}

\icmlcorrespondingauthor{Prasang Gupta}{prasang.gupta@pwc.com}
\icmlcorrespondingauthor{Shaz Hoda}{shaz.hoda@pwc.com}

\icmlkeywords{Machine Learning, ICML, Investment, RL, Deep Learning, Evolutionary Strategy, GA}

\vskip 0.3in
]



\printAffiliationsAndNotice{\icmlEqualContribution} 

\begin{abstract}
    Machine learning driven trading strategies have garnered a lot of interest over the past few years. There is, however, limited consensus on the ideal approach for the development of such trading strategies. Further, most literature has focused on trading strategies for short-term trading, with little or no focus on strategies that attempt to build long-term wealth. Our paper proposes a new approach for developing long-term investment strategies using an ensemble of evolutionary algorithms and a deep learning model by taking a series of short-term purchase decisions. Our methodology focuses on building long-term wealth by improving systematic investment planning (SIP) decisions on Exchange Traded Funds (ETF) over a period of time. We provide empirical evidence of superior performance (around 1\% higher returns) using our ensemble approach as compared to the traditional daily systematic investment practice on a given ETF. Our results are based on live trading decisions made by our algorithm and executed on the Robinhood trading platform.
\end{abstract}

\section{Introduction}
\label{submission}

The stock market plays plays a crucial role in wealth building for retail investors. With modernization of financial and information systems, the large amount of information available for a trader has made analysis of a financial asset prohibitive using simple analytical methods. Due to the widespread availability of information and the complexity of deciphering patterns from them, a large number of scientific papers have been published to investigate computer driven techniques for solving financial market problems \citep{cavalcante2016computational}, implementing a machine learning driven trading strategy involves developing procedures to: 1) predict the movement of prices, and 2) make decisions at the optimal time in an attempt to maximize returns based on the predicted movement. Many studies tackle both problems within a single approach \citep{lv2019empirical}. A number of approaches using machine learning have been explored \citep{boyd2017multi, gerlein2016evaluating, dash2016hybrid, huang2019automated}. Some approaches use reinforcement learning for developing short-term trading strategies \citep{deng2016deep, xiong2018practical,  li2019deep, zhang2020deep}, while others use genetic algorithms \citep{mendes2012forex, chen2019effective}. A consolidated survey paper with extensive comparisons between the different modeling techniques is given by \citep{lv2019empirical}. The most comprehensive review of machine learning experiments in equity investments is given by \cite{buczynski2021review}. They conduct a detailed review of 27 academic experiments spanning over two decades and contrast them with real-life examples of machine learning-driven funds. They also provide recommendations on how to approach future experiments.

A further observation on these studies is that they have focused mostly on the performance of short-term buy/sell decisions. The literature is lacking long-term, multi-year, investing and wealth building techniques that use machine learning. While some studies address long-term performance, they use rule-driven and technical analysis based trading strategies \citep{giamouridis2017systematic}.

The focus of this paper is to propose a new method using an ensemble of evolutionary strategies and deep learning as an alternative to reinforcement learning. Further, this paper addresses a gap in the literature by focusing on training an algorithm that helps average retail investors build wealth over time by exploiting short-term volatility in Exchange Traded Funds (ETFs). Due to short-term volatility, such as decline in prices, an opportunity exists to buy ETF when it's at its lowest. This is in contrast to short-term high frequency trading algorithms that the current literature focuses on. Finally, we address some of the recommendations laid out by \cite{buczynski2021review} to increase confidence in our results and encourage practical adoption of machine learning models in finance.

The main contributions of this paper are as follows.
\begin{enumerate}
    \item We introduce algorithm \ref{algo:GADLE algorithm} that uses an ensemble of a genetic algorithm and a deep learning model and serves as an alternative to traditional reinforcement learning techniques. We call this algorithm GADLE (Genetic Algorithm \& Deep Learning Ensemble) for short. 
    \item We exploit short-term volatility in ETFs to make superior daily investment decisions over a traditional daily systematic investment plan (SIP). 
    \item We introduce a concept we call 'contextual scaling', that can help make long history of data comparable for episodic tasks of reinforcement learning
    \item We address some of the key concerns highlighted in the paper by \cite{buczynski2021review} such as problem of 'cherry-picking', production testing of models
\end{enumerate}

\begin{algorithm}[t]
\SetAlgoLined
\caption{GADLE Algorithm}
\label{algo:GADLE algorithm}
$\mathbf{m}$, maximum number of iterations\\
$\mathbf{\mathcal{N}_{pop}} = 100$, population size for GA\\
$\mathbf{\mathcal{P}_{cross}} = 0.4$, crossover probability for GA\\
$\mathbf{\mathcal{P}_{mut}} = 0.2$, mutation probability for GA\\
\While{not all episodes are solved}{
    $i \gets 0$\\
    $\mathbf{\mathcal{S}} = \{\vec{a}_{1}, \vec{a}_{2}, \vec{a}_{3},\:...\:, \vec{a}_{\mathcal{N}_{pop}}\}$, randomly generated action vectors\\
    Calculate the loss $\mathcal{L}_{i}$ for each $\vec{a}_{i}$\\
    \While{$i <= \mathbf{m}$ \textbf{or} \textit{max iterations without improvement}}{
        $\mathbf{\mathcal{S}}$, set of top performing action vectors $\vec{a}_{i}$'s from previous iteration\\
        Perform crossover on $\mathbf{\mathcal{S}}$ based on $\mathbf{\mathcal{P}_{cross}}$\\
        Perform mutation on $\mathbf{\mathcal{S}}$ based on $\mathbf{\mathcal{P}_{mut}}$\\
        Evaluate action vectors by calculating loss $\mathcal{L}$\\
        $i \gets i+1$
    }
    Store optimum action vector for this episode $\vec{a}^{*}$ in $\mathbf{\mathcal{A}}$\\
    Store scaled historical prices $\vec{p}$, moving averages $\bar{p}$ and price movements $\vec{p}_{change}$ for this episode in $\mathbf{\mathcal{X}}$
}
Train MLP network with $\mathbf{\mathcal{X}}$ as independent variable and $\mathbf{\mathcal{A}}$ as dependent variable
\end{algorithm}
\vskip -0.25in

\subsection{Problem Statement}

Consider an investor that wants to invest in a particular ETF over a period of time to accumulate wealth. Once an ETF is chosen, the investor can decide whether or not to continue to invest in this ETF on a periodic basis to build wealth. An investor who chooses to invest in an ETF a fixed amount on a daily basis is said to subscribe to a daily Systematic Investment Plan (SIP). By investing regularly (daily/weekly/monthly) in an SIP, the investor is less susceptible to the short term fluctuations of the market and in the long run, can make annualized returns in the range of 8-11\% \footnote{https://www.investopedia.com/ask/answers/042415/what-average-annual-return-sp-500.asp Accessed May 28, 2020.}. Our objective is to, exploit the short-term volatility instead of evening it out and hence improve on this return by 10\% on an annual basis (8.8-12.1\%), which would provide tremendous compounding effects in the long-term.

We found very limited literature that focuses on wealth building over time by making marginally better daily buying decisions over a lifetime of investing \citep{wang2021robo, philps2018continual}. However, these models still rely on either a 12 month evaluation window to train agents or are more focused on risk profile assessment and portfolio balancing over time. 

To solve the problem of continuous, marginally superior investment decisions in the long run, we break down our infinite investment horizon into a series of finite windows of 30 trading days in which an investor can decide on each day to invest either twice, or nothing at all in the ETF of their choice. We choose a 30 trading day decision window because it is long enough to offer some volatility in the ETF's price, while short enough to find optimal investing days without significant computational overhead. We allow the algorithm more flexibility than conventional daily investment strategies with the option to either not buy or buy twice in the day. By exploiting day-to-day volatility, it improves performance over conventional systematic investment plans (SIPs) which are forced to buy daily.  

Further, we assume that the investor has enough liquidity at any point in time to make double their regular target monthly investment. This is an important liquidity assumption and effectively allows our algorithm to make twice the investment, if required, that is targeted by the investor at any given point of time. For example, if the ETF is on a constant decline for 30 trading days, this assumption gives our algorithm the flexibility to purchase 60 units of the ETF as opposed to only 30 units for an investor who is limited by 30 days of liquidity at any point.

If we can train our algorithm to make an equal number of buy twice / not-buy actions (denoted by $a=2$ and $a=0$) on average, this would result in a net investment - over multiple 30 day trading periods - that is very close to the daily buying investment ($a =1$) strategy, per the law of large numbers. That is, the expected number of purchase actions ($N_{a}$) would converge to the same number under both strategies as the number of periods becomes large. 
\setlength{\abovedisplayshortskip}{1pt}
\setlength{\belowdisplayshortskip}{1pt}
\vskip -.1in

\begin{equation}
    \mathbb{E}[N_{a=1}] = \mathbb{E}[N_{a=2} + N_{a=0}]
\end{equation}

Our objective is to train a model to learn a strategy that can make optimal investment decisions over a 30 trading day period such that the investor is better off using the agent's decisions rather than passively investing a fixed quantity daily.

To train the agent to take optimal decisions on a daily basis, we can potentially apply a traditional reinforcement learning (RL) approach in which the agent observes the market condition (the state), makes a decision (don’t buy or buy twice), observes the reward (or loss), either per period or after many years of investing, and learns the optimal parameters of the network that estimate the value of a state or the value of an action as explained in classic reinforcement learning principles by \citep{sutton2018reinforcement}.

This problem in the above form is similar to training an agent to play Atari games such as Pong, Breakout, Space Invaders \citep{mnih2013playing}, or teaching an agent to walk \citep{haarnoja2018learning}, with the simplifying assumption that an average retail investor on their own can't influence or sway the outcome of the stock market. That is to say, the environment is exogenous; all the retail investor can do is observe the state of the market and take an action.

A significant amount of work has been done in the domain of RL with Deep-Q Learning \citep{mnih2013playing} and the more state of art Actor-Critic \citep{mnih2016asynchronous} and we use these to create benchmarks against which we compare our proposed approach. As an alternative to the two RL approaches in DQN algorithm \ref{algo:DQN Algorithm} and Actor-Critic algorithm \ref{algo:Modified Actor-Critic} that are discussed in detail in appendix \ref{appendix:DQN and actor}, we present our own algorithm - Genetic Algorithm \& Deep Learning Ensemble (GADLE). 

Our GADLE approach is primarily inspired by 2 different and unrelated papers - \citep{salimans2017evolution} and \cite{cobbe2021phasic}. These papers have different contributions to the literature which we draw upon for our approach.

The paper \textit{Evolutionary Strategies (ES) as an alternative to Reinforcement Learning} suggests ES as an alternative optimization algorithm for finding good neural network weights as opposed to using gradient descent for RL problems. In their paper, each neural network is independently created and exchanges only a small amount of information each generation. \citep{salimans2017evolution}. The paper demonstrates the advantages of using ES over RL:
\begin{itemize}
\item The code is 2-3 times faster in terms of overall runtime
\item ES is highly parallelizable
\item ES has fewer hyperparameters
\item ES is a more attractive choice for gradient estimation when number of time steps in an episode is long
\end{itemize}
While solving for weights in the neural network is more rapid using an ES approach, the value function and policy function are still tightly coupled. That is, the network is trying to find the optimal answer in parallel to finding the optimal policy that leads to the answer.

The papers on decoupling value and policy in reinforcement learning by \cite{cobbe2021phasic} and more recently by \cite{raileanu2021decoupling} propose a solution in which they distribute the internal neural architecture of Actor-critic network. The overall network, however in the decoupled approach still has to exchange information every episode, which does not offer any speed up, although it makes the learning more generalizable. 

Our GADLE algorithm combines the ideas of \cite{salimans2017evolution} and \cite{cobbe2021phasic} by using Genetic Algorithm to solve for optimal actions (to get a tremendous boost in training speed) and by completely decoupling value \& policy by training an independent neural network based on optimal actions (to further reduce the workload). We do not function within the domains of a single RL algorithm and distribute the workload across GA and DL, both of which are separate methods.
Our algorithm has further advantages over the approach proposed by \citep{salimans2017evolution}, in which the solving for optimal weights of the neural network still happens in parallel to finding the optimal solution to the overall problem. In our proposed method - the two processes are decoupled completely. Evolutionary Strategy solves each environment (while the overall policy that lead to the solution is still unknown). Once multiple such environment is solved (target is known), a neural network is trained independently to learn the optimal policy. This is much simpler process, effectively two independent programs - one using Evolutionary Strategy to solve the environment and the other using neural network to learn the policy that would give the same solution. We show that our model can run with high speed on a basic 16 core machine, while the approach by \citep{salimans2017evolution} uses heavy compute 1,000+ cores to achieve the speed up they report in their paper.

We present the pseudo code for our GADLE algorithm \ref{algo:GADLE algorithm} to learn optimal actions and subsequently train a neural network on the optimal actions. 
In section 2 of the we discuss in detail the data and methodology for GADLE and create benchmarks for comparison against DQN and Actor-critic in section 3. In the results section 4, we compare the performance of our GADLE algorithm to that of DQN and Actor-Critic network in terms of ease of architecture, time to train, difficulty in finding optimal parameters, consistency and actual performance. 

Our approach, substantially simplifies traditional ways of solving this Markov Decision Process (MDP) by decoupling the learning of policy networks from solving the optimal action for each state.  This practice not only reduces code complexity substantially (as we show later in section 4), but also introduces a new framework altogether for solving MDPs that builds upon the work done by \citep{salimans2017evolution} and  \cite{cobbe2021phasic}. Finally, we show that our model trained via this approach performs well in practice by implementing our code for production testing.

\section{Data and Methodology for Proposed GADLE Algorithm}

Our overall approach can be broken down into a series of steps that we cover in detail in this section. As mentioned in the previous section, we solve each episode independently based on the simplifying assumption that rewards are observed in each episode and is not a one time/rare event. The steps we follow are laid out in the sub-sections below.
 
\subsection{Creating environment simulation}

Previous works have detailed procedures for simulating stock market price movements \cite{cui2012intelligent,souissi2018multi, byrd2019abides}. Instead of creating a simulator that mimics the movement of the stock market, we create an environment sampler that creates episodic samples based on actual stock market data. 
We retrieve data on the Vanguard Total Stock Market Index (VTI) from Yahoo Finance to create 30 trading day samples of episodes without replacement (2 samples can't be exactly the same, though they can have some overlap). We train using the period 2000-1-1 to 2019-12-31 and we leave the period of 2020-01-01 to 2020-12-31 for back-testing. We assume that the average retail investor is not placing trades large enough to influence price movements of the index fund. As such, our environment samples can be treated as fixed.

\subsection{Creating episodes}

A common issue encountered while creating samples is the change in price scale over time. Since stock price in 2000 will be very low as compared to that in 2020, using the same scaler for both years would lead to training a scaler that treats price increase over time as a feature itself. To address this, we introduce a concept that we call ‘contextual scaling’. This concept becomes pertinent as for long trading horizons. We do contextual scaling by sampling 60 days of data from the 20 year history of the fund for each episode. We then use the first 30 of 60 days to fit the scaler and then apply the scaler to the next 30 days. This way, each episode is scaled to the context of the 30 days prior to it and we avoid the problem of change in scale across episodes overtime.

We run the sampler 4,245 times to create 4,245 independent episodes, each ‘contextually scaled’, to make it comparable to other episodes for training.

\subsection{Defining a loss function}

For each episode, the agent has 30 trading days in which the agent must decide whether to buy twice or not not buy. As mentioned, we solve each episode independently using ES with a loss function that ideally captures the benefits of buying twice versus not. Our loss function is defined as follows:
\vskip -0.25in
\begin{align}
  \mathcal{L} = \min_{a} \dfrac{\left(\dfrac{\vec{p} \cdot \vec{a}}{N_{a=2}} - \overline{p}\right)}{\overline{p}}*\left(2 N_{a=2} \right) + {\left(1 - \frac{N_{a=2}}{15}\right)}^{2}
\end{align}
\vskip -0.1in
   
The loss function that will be used to solve each episode has 2 major components. The first component, $(\frac{\vec{p} \cdot \vec{a}}{N_{a=2}} - \overline{p})*(2 N_{a=2})$  captures the return over a daily investment plan by subtracting the daily average price $(\overline{p})$ from the agent average price $(\frac{\vec{p}\cdot \vec{a}}{N_{a=2}})$ and multiplies it with the number of purchases. Here $\vec{p}$ is the price vector and $\vec{a}$ is the action vector for the 30 day window. The second component, $(1 - \frac{N_{a=2}}{15})^{2}$, behaves like a regularization factor and rewards/penalizes the agent for infrequent or overly frequent purchasing decisions. We square the second term for higher penalties on larger deviations such that small over-buy and under-buy decisions are not penalised as heavily as larger deviations. When the agent purchases (twice) for 15 of 30 days this factor is 0. If the agent purchases 0 of the 30 days then this factor adds to the loss a factor of 1. Finally, when the agent purchases (twice) on all 30 days, the first component is 0 and the return is taken as 1.

\subsection{Solving each episode using ES}

Given our loss function, we use ES to find the optimal solution for each episode (optimal sequence of actions that minimizes the loss function). We use genetic algorithms (GA) as a mechanism for our ES. 
We use a classical GA \citep{holland1975adaptation}, where each chromosome is a string of bytes. In our case, each episode becomes a chromosome of length 30, and each decision that can be taken by the agent (1 - buy twice, 0 - don’t buy), becomes a gene within the chromosome. Each chromosome (vector of length 30) represents a single possible solution to our optimization problem and the population of chromosomes is the set of candidates for the optimum solution. The hyperparameters used to configure our GA can be found in Appendix \ref{appendix:GA param and use}. 

For each solved episode, the action decisions is stored. This exercise is repeated for the entire period for which the ETF existed to generate 4,245 independent episodes (maximum possible unique episodes, given the history).

\subsection{Overview of data generated}

In this section we present an overview of the distribution of returns and actions for the solved episodes. The objective gauge the performance of actions across 4,245 episodes. Another objective is to see any evident biases in action behavior or returns that we would not want our agent to have. A detailed version of this section is presented in Appendix \ref{appendix:GADLE data review}.

\vskip -.1in
\begin{figure}[ht]
\caption{Joint distribution of returns and purchase decisions}
\begin{center}
\centerline{\includegraphics[width=.3\textwidth]{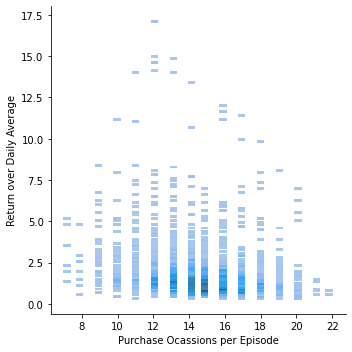}}
\label{fig:2 joint distribution across episode}
\end{center}
\vskip -.35in
\end{figure}

Figure \ref{fig:2 joint distribution across episode} shows the distribution of the agent's return over daily average prices for the episode. 
We can see from the plot that we are able to solve for each episode independently with our loss function, yet able to produce results that, on average, would be able to maintain a portfolio size equal to the daily purchase behavior, while still attaining a return that is higher than the daily average strategy. Our next step is to determine if we can train a neural network model (a policy function) that can learn the behavior of this agent across episodes.

\subsection{Fitting the Deep Learning Model on Solved Episodes}

As mentioned before, we use a GA to independently solve each episode and then train a neural network model to learn the optimal strategies across these episodes. The decoupling of solving for optimal action while training the neural network to take optimal action greatly simplifies the process of solving this MDP. 



We train a deep learning (DL) model, where the input is a sequence of prices $\vec{p}$ and predict the actions $\vec{a}$, that we obtained through from the GA for all 4,245 episodes. As mentioned, since each episode's prices are already scaled independently using ‘contextual scaling’, we no longer need to worry about making episodes sampled across years comparable. We simply create features before scaling, which capture the price movement for each state within each episode. We feed these features into a simple DL model along with state specific features, such as the number of purchases made up to time $t-1$ in the episode. We train a feed forward neural network with 6 layers and ReLU activation function at each layer. We divide our 4,245 episodes into 3,745 episodes for training and validation and hold out 500 episodes for the test set.

Our basic model achieves an accuracy of 93.37\% in predicting the optimal agent action sequence obtained earlier through the GA on the 33\% of our 3,745 episodes reserved for validation. Our model returns an accuracy of 93.92\% on the test dataset (500 episode holdout). These accuracies show that our neural network is able to learn an overall policy function that closely resembles the actions taken by the GA to solve each episode. A more detailed assessment of our decoupled model can be found in Appendix \ref{appendix:additional NN details}. 

\begin{equation}
    \text{RoD} = \left(1-\dfrac{\left(\dfrac{\vec{p} \cdot \vec{a}}{N_{a=2}}\right)}{\overline{p}}\right) * 100
    \label{eq:rod}
\end{equation}

\begin{equation}
    \text{PCoD} = 2N_{a=2} - 30
    \label{eq:pcod}
\end{equation}

We finally perform a final out of sample validation (back-testing), before putting our model into production on data from 2020-1-1 to 2020-12-31. This strategy is evaluated against a daily investment strategy on the ETF VTI. We have used 2 metrics to evaluate our strategy. One metric is RoD (Return over Daily), shown in equation \ref{eq:rod}, which quantifies the percentage return of our strategy based on price differences over daily investment. The second metric that we have used is PCoD (Purchase Count over Daily), shown in equation \ref{eq:pcod} which is the difference in the stock purchase amounts made by our agent over daily (30 for a month).

The results from back-testing are shown in Table \ref{tab:esresults}. We see the superior performance of our agent as compared to a simple daily investment strategy by a significant 1.26\% margin. The agent, however, makes 230 unit purchases versus a daily purchase of 240. We treat this as a tolerable behavior and reserve this as a problem that can be solved in further work by fine-tuning the loss function.

\begin{table*}[t]
\centering
\caption{GADLE Agent return and purchase decisions over daily SIP}
\label{tab:esresults}
\begin{small}
\begin{sc}
\begin{tabular}{lccccccc}
\toprule
{2020 Periods} & \multicolumn{3}{c}{Average Price} & & \multicolumn{3}{c}{No. of purchases} \\ \cline{2-4} \cline{6-8}
& Agent & Daily & RoD & & Agent & Daily & PCoD \\
\midrule
Jan 2nd - Feb 13th  & 160.86 & 161.98 &  0.68 \% & & 30 & 30 &  0 \\
Feb 14th - Mar 27th & 129.77 & 139.38 &  6.89 \% & & 38 & 30 &  8 \\
Mar 30th - May 11th & 137.74 & 135.13 & -1.93 \% & & 2  & 30 & -28 \\
May 12th - Jun 23rd & 146.47 & 149.88 &  2.27 \% & & 34 & 30 &  4 \\
Jun 24th - Aug 5th  & 154.56 & 157.93 &  2.13 \% & & 26 & 30 & -4 \\
Aug 6th - Sep 17th  & 167.44 & 169.10 &  0.98 \% & & 42 & 30 &  12 \\
Sep 18th - Oct 29th & 168.30 & 169.61 &  0.76 \% & & 44 & 30 &  14 \\
Oct 30th - Dec 11th & 172.05 & 180.83 &  4.85 \% & & 14 & 30 & -16 \\
\midrule
Overall             & 155.99 & 157.98 &  1.26 \% & & 230 & 240 & -10 \\
\bottomrule
\end{tabular}
\end{sc}
\end{small}
\end{table*}
\vskip -0.25in
\section{DQN and Actor Critic Benchmark Creation}

Before we set out to test our proposed algorithm in production environment, we also benchmark the performance of our proposed algorithm against Actor-Critic and DQN for completeness. We use the same data as described the previous section to create benchmark performance of the algorithm against the two most common RL algorithms.

\begin{equation}
    \mathcal{EP}_{i} = \dfrac{i}{30} \;,\; i \in \{1,2, ... ,30\}
    \label{eq:episodeprogress}
\end{equation}

\begin{equation}
    \mathcal{BR}_{i} = \dfrac{\sum\limits_{k=1}^{i} N_{a=2, k}}{i} \;,\; i \in \{1,2, ... ,30\}
    \label{eq:buyratio}
\end{equation}

There is a fundamental difference between how this problem is solved by our proposed GADLE method and the RL method. In the GADLE method, we are solving for a month at once. Hence, actions for the month are decided at a go and the reward function and further iterations are done at a monthly level. Implementing this one-to-one in RL is not feasible due to the huge action space involved ($2^{30}$ possible actions). Hence, we converted this to an episodic RL problem with episodes spanning each month and solving for one day at a time (in a single timestamp). This problem formulation has one shortcoming. Since we are solving for each day at a time, the agent currently doesn't have any information to use to control its buying patterns. Hence, we added 2 additional features to the state space. These 2 features are Episode Progress ($\mathcal{EP}$) and Buy Ratio ($\mathcal{BR}$) defined as in Equation \ref{eq:episodeprogress} and \ref{eq:buyratio}. 
Another fundamental difference between the RL methods and the GADLE is the problem formulation. While the GADLE method is a minimisation problem based on the Loss function defined earlier, the RL methods are a maximisation problem for the reward. Hence, we have modelled the reward for these methods as the negative of the loss function to keep them comparable to the results of the GADLE method.

Having discussed the general changes done to convert the problem into an RL problem, details of DQN agent training and Actor-Critic agent training is presented in Appendix E.

\section{Findings and Results}

\subsection{Comparison of Algorithms}

We present in this section a comparison of performance of DQN, Actor-Critic Network and our GADLE Approach across the four parameters: 1) Time taken to train; 2) Sensitivity to hyper-parameters; 3) Consistency with changes in random seed; 4) Qualitative Assessment

\subsubsection{Time to train}

Both the GADLE approach and the RL methods have a NN architecture embedded into the overall architecture and hence, need some time to train before evaluation. We have used the same compute power to train all of these algorithms but capped the training time to 24 hours. We ran all of these on a machine with 16 Core Intel CPU and an NVIDIA P100 GPU. Comparison for time taken to train for different algorithms can be seen in Table \ref{tab:timetotrain}.

\begin{table}[t]
\centering
\caption{Time to train comparison for different methods}
\label{tab:timetotrain}
\begin{small}
\begin{sc}
\begin{tabular}{lcccr}
\toprule
Method & Total Time & Episodes & Speed (ep/hr)\\
\midrule
GADLE & 33 min & 4245 & 7718 \\
DQN & 24 hrs & 1440 & 60 \\
Actor-Critic & 24 hrs & 288000 & 12000 \\
\bottomrule
\end{tabular}
\end{sc}
\end{small}
\end{table}

\begin{table*}[ht]
\centering
\caption{Hyperparameter sensitivity of algorithms}
\label{tab:sensitivityresultssummary}
\begin{small}
\begin{sc}
\begin{tabular}{llcccc}
Algorithm & Metric & Agent Avg. & RoD. & PCoD & Failed \\
\midrule
{GADLE: 10 tests} & Baseline & 155.99 & 1.26\% & -1.25 &{0 / 10} \\
& Avg. across hyperparamters & 156.33 & 1.05\% & -1.98  \\
& Avg. absolute deviation from baseline & 0.51 & 0.33\% & 1.42   \\
\cmidrule{2-5}
{Actor-Critic: 10 tests} & Baseline & 155.80 & 1.38\% & -1.25 &{2 / 10} \\
& Avg. across hyperparamters & 158.35 & -0.23\% & -2.81  \\
& Avg. absolute deviation from baseline & 2.77 & 1.75\% & 4.61    \\
\bottomrule
\end{tabular}
\end{sc}
\end{small}
\end{table*}

\begin{table*}[t]
\centering
\caption{Consistency comparison for all algorithms}
\label{tab:consistencyresults}
\begin{small}
\begin{sc}
\begin{tabular}{lccccc}
\toprule
Method & Mean RoD & Std. Dev. RoD & Mean PCoD & Std. Dev. PCoD & Fail \%\\
\midrule
GADLE & 1.52 & 0.42 & -2.01 &  1.56 & 0 \% \\
DQN            & 0.53 & 3.23 & -0.17 & 13.53 & 35 \% \\
Actor-Critic   & 0.79 & 1.78 & -1.34 & 10.51 & 12.5 \% \\
\bottomrule
\end{tabular}
\end{sc}
\end{small}
\end{table*}

\begin{table*}[ht]
\centering
\caption{Agent return and purchase decisions over daily SIP}
\label{tab:agent_prod_resukts}
\begin{small}
\begin{sc}
\begin{tabular}{lccccccc}
\toprule
{2021 Periods} & \multicolumn{3}{c}{Average Price} & & \multicolumn{3}{c}{No. of purchases} \\ \cline{2-4} \cline{6-8}
& Agent & Daily & RoD & & Agent & Daily & PCoD \\
\midrule
Jan 4 - Feb 15  & 197.17 & 200.16 & 1.49\% & & 20 & 30 & -10  \\
Feb 16 - Mar 29 & 203.64 & 204.58 & 0.45\% & & 42 & 30 & 12  \\
Mar 30 - May 12 & 211.36 & 214.87 & 1.63\% & & 18 & 30 & -12 \\
May 13 - Jun 24  & 216.23 & 217.97 & 0.79\% & & 22 & 30 & -8  \\
Jun 25 - Aug 08 & 218.74 & 225.19 & 2.83\% & & 2 & 30 & -28  \\
Aug 09 - Sep 20 & 228.79 & 230.50 & 0.74\% & & 32 & 30 & 2 \\
Sep 21 - Nov 01  & 228.93 & 229.52 & 0.26\% & & 48 & 30 & 18  \\
Nov 02 - Dec 14 & 236.00 & 239.01 & 1.26\% & & 22 & 30 & -8  \\
\midrule
Overall               & 218.44 & 220.22 & 0.81\% & & 206 & 240 & -34\\ 
\bottomrule
\end{tabular}
\end{sc}
\end{small}
\end{table*}

It can be seen that the total time to train for the proposed GADLE approach is just 33 minutes as compared to the 24 hours (maximum permitted time) for either of the RL methods. One thing to note is that we have ignored the time to prepare data for training and other steps as it is common across all the 3 different algorithms with very little variation. It can also be seen that the speed (episode / hour) for the DQN algorithm is worst and it is best for Actor-Critic. However, the episodes needed for the Actor-Critic algorithm - due to the inherent nature of RL methods - to yield good results are much larger than the episodes needed for the GADLE approach. Hence, this in turn yields that the GADLE approach trains much faster.

\subsubsection{Sensitivity}

Another difference between our proposed GADLE method and the RL based methods is the difficulty of finding optimal parameters. The GADLE method requires parameters for Genetic Algorithm (that are listed in Table \ref{tab:ga_parameters} in appendix \ref{appendix:GA param and use}) while the RL based methods require parameters for Learning Rate, Decay and Discount Factor (highlighted in Table \ref{tab:a2c-parameters} and Table \ref{tab:dqn-parameters} in appendix \ref{appendix:DQN and actor}). However, we observed that finding the optimal values for GADLE parameters was much easier than RL parameters.

To highlight this, we performed a sensitivity analysis on some hyperparameters and compared the performance of the methods. If the model performance does not change much with slight change in hyperparameters, it highlights the stability of the model with respect to its hyperparameters. Conversely, we can say that finding optimal parameters is much easier and achievable in models with stable hyperparameter dependence.
To conduct sensitivity, we changed several hyperparameters for GADLE, DQN and Actor-critic RL methods by $\pm$ 20\%. After changing these hyperparameters models were retrained and evaluated on 2020 data. A detailed sensitivity analysis can be found in appendix \ref{appendix:sensitivity runs}. A summary of the same is presented in table \ref{tab:sensitivityresultssummary}.
Looking at the summary table for GADLE method, the Agent Average price deviation comes out to be 0.51 and that of RoD (return over daily) and PCoD (purchase count over daily averaged over a month) comes out to be 0.33 \% and 1.42 respectively, with no failed runs. These deviations are very low as compared to Actor-critic method, for which the RoD and PCoD variation is much higher at 1.75 \% and 4.61 respectively, with 2 failed runs.
It is immediately clear that performance of all variations are very comparable to that of our original run for GADLE while the same can not be said about Actor-critic. We didn't include sensitivity results for the DQN method as it was so unstable, that even a slight change in its parameters lead to the run crashing and the model obtained either purchased nothing or everything.


\subsubsection{Consistency}
\label{sec:consistency}

We discuss here the consistency of the performance of the models. We show this by running the same code for multiple iterations using different random seeds and then highlighting the variance in the performance metrics. We selected the same two metrics for evaluating the performance as described in Equations \ref{eq:rod} and \ref{eq:pcod} i.e. Agent RoD (return over daily) and Agent Monthly PCoD (purchase count over daily averaged over a month). The former metric highlights the efficacy of the agent to select the best buy options and the second metric helps keep a track of any under-purchasing or over-purchasing over the daily agent.


We ran all the different methods 40 times with a different random seed to check their consistency and report RoD and PCoD scores when evaluated on the 2020 data.

We can see the variation of performance for all three methods in table \ref{tab:consistencyresults}. The GADLE method was found to be very robust in terms of returns, with all 40 iterations performing better than the daily agent with a mean RoD of 1.52\% and a standard deviation of 0.42\%.  However, there is a slight under-purchasing trend that can be observed in some of the cases, but it was observed that more than 50 \% of the iterations were within the $\pm$ 2 PCoD region.
As compared to this, the performance of Actor-critic method was less promising. 5 of the total 40 runs were marked unsuccessful as those runs had below par reward progressions as compared to the others. For the remaining 35 runs, roughly 25 \% of the runs were found to be worse than the daily agent and only about 50 \% of the runs were found to yield RoD value greater than 1 \%. Further, a mean RoD of 0.79\% and a high standard deviation of 1.78\%, the performance was found to be vastly inferior to GADLE method.
Due to the poor performance of DQN in previous section, we did not feel the need to elaborate the results further.
Further details and comparisons around the training and RoD, PCoD distributions can be found in appendix \ref{appendix:consistency runs}

\subsubsection{Qualitative Assessment}

We qualitatively compared the algorithms using responses of a few data scientists who had no previous experience in any of these methods. It was the general consensus that our proposed GADLE method was much more straightforward and easier to understand than the other RL based methods (DQN and Actor-Critic). It was also found that the effort in writing a working code was much more in RL based methods than the GADLE approach. We calculated the approximate lines of code that was written to implement the solution end-to-end, assuming proper libraries are imported and after discounting print statements, import commands, docstrings etc. 
The number of lines of code for the GADLE approach are 30-45\% less than the RL-based approaches. One of the reasons for that is the GADLE approach based on GA doesn't need an environment class to be defined to get the actions and learn the optimal ones, which RL methods do. Interacting with the environment class, defining the learning algorithm with support for interaction as well as the large number of hyper parameters, all increase the complexity of the RL based methods over the GADLE approach.

\subsection{Experiment in Production Environment}

Finally, we retrain our agent with all available data from 2000-1-1 to 2020-12-31 and put our model to production, using one of the author's Robinhood trading account and on a cloud virtual machine. Our model runs at 12PM EST everyday and lets the agent take an action of buy twice or hold starting January 2$^{nd}$, 2021. These results from live agent actions are presented in table \ref{tab:agent_prod_resukts}. The architecture for executing our algorithm is presented in appendix \ref{appendix:production_architecture}.

Table \ref{tab:agent_prod_resukts} shows that our agent once trained on data from 2001-1-1 to 2020-12-31 for the ETF VTI, performs very well against a daily SIP strategy. We can see that the agent makes buy/hold decisions evenly, based on its trained experience, when it expects the market to be under-priced. For each 30 day trading window during 2021, the agent outperforms a daily purchase strategy. Overall the agent improves the daily return by 0.81\%. We do observe slight under purchase behavior of our agent. However, given that our agent learned from episodes, which on average made 15 (buy twice) purchase decisions during a 30 day window, our agent should in the long run converge to a number closer to daily SIP trade counts. We discuss some potential reasons for under-purchase in the next section. 



\section{Discussion}
 
 Our paper introduced a new algorithm, GADLE, that serves as an alternative to traditional RL techniques. We showed how a long-term wealth building task can be broken down into smaller 30 day episodic tasks and how an agent can be trained on 20 years of historic data broken down into distinct episodes scaled by the context. Our GADLE algorithm demonstrated a 1.26\% out-of-sample tests. Our agent also performed well in live production environment as well, with an average return that was 0.81\% higher than the traditional SIP. We alleviated some of the key concerns in the domain of machine learning and investing research raised by \citep{buczynski2021review} by carrying out multiple consistency checks for our model and deploying it in production environment.
 We did see slight decrease in performance in production environment due to factors such as a) our model was trained on average price for the day, while applied on a single price point during the day; b) we used a python package (yfinance) to fetch stock market data for the ETF that we found to be inconsistent such that the actual price during the day can be different for different pulls of the same data; c) we used a vanilla neural network that achieved a 93\% accuracy, a better architecture / model would in theory yield results that are superior; d) and finally, the agent perhaps expects a market correction and hence is under-purchasing continuously in anticipation of a decline.
 Future work would include addressing some of the limitations above and improve feature engineering and model architectures to improve the accuracy of our neural network. There is scope to use a superior loss function that improves returns by enabling the agent to learn better behavior. Another avenue to explore is to allow the agent to periodically sell holdings to minimize taxes on returns. Future work can also incorporate the cost of borrowing, which can be added to the loss function instead of assuming investor liquidity.
 

\newpage
\bibliography{bibliography}
\bibliographystyle{icml2022}

\newpage
\appendix
\onecolumn

\section{Genetic Algorithm Hyperparameters and Usage}
\label{appendix:GA param and use}

Our GA part of the GADLE algorithm is configured with the following hyperparameters:

\begin{table}[htbp]
\centering
\caption{Parameters for genetic algorithm}
\label{tab:ga_parameters}
\begin{small}
\begin{sc}
\begin{tabular}{llc}
\toprule
Parameter & Definition & Value \\
\midrule
Number of iterations & Maximum number of tries                     & 200            \\ 
Population size      & Number of trial solutions in each iteration & 100            \\ 
Mutation probability  & Chance of each gene to be replaced by a random value                   & 0.2 \\ 
Elite ratio           & Proportion of best performing genes carried over to the next iteration & 0.3 \\ 
Crossover probability & Chance of existing solution passing its genome to new solutions        & 0.4 \\ 
Parents proportion    & Proportion of population filled by members of previous generation      & 0.3 \\ 
Stopping criteria     & Number of iterations without any improvement in loss                   & 30  \\ 
\bottomrule
\end{tabular}
\end{sc}
\end{small}
\end{table}

Readers should note that using GA to solve for training a DL model can only be applied with the following assumptions:
\begin{itemize}
\item The game can be broken down into independent windows of constant length (episodes).
\item The game must continue for a sufficiently large number of episodes.
\item Rewards are observed in each episode and not a one time or rare event (for example, this method can’t be applied to a game of chess where reward is observed only when the game ends, unless each action on the board has a reward associated with it)
\item A machine learning model can be trained to predict the ideal outcome of each episode with a high degree of accuracy.
\end{itemize}

\section{Detailed review of GADLE model data (GA Part of GADLE)}
\label{appendix:GADLE data review}
\begin{figure}[htbp]
    \centering
    \begin{subfigure}[t]{0.35\textwidth}
        \centering
        \includegraphics[width=\textwidth]{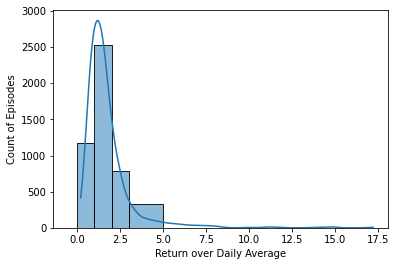}
        \caption{Distribution of returns across episodes}
        \label{fig:ga_distribution_returns}
    \end{subfigure}
    \hfill
    \begin{subfigure}[t]{0.35\textwidth}
        \centering
        \includegraphics[width=\textwidth]{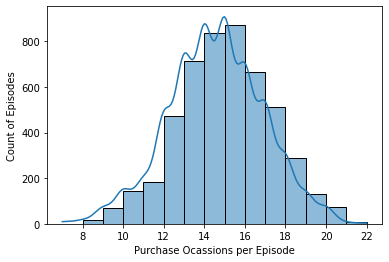}
        \caption{Distribution of purchase decisions across episodes}
        \label{fig:ga_distribution_purchasedecision}
    \end{subfigure}
    \hfill
    \begin{subfigure}[t]{0.25\textwidth}
        \centering
        \includegraphics[width=\textwidth]{figures/gadle_joint_ep.png}
        \caption{Joint distribution of returns and purchase decisions}
        \label{fig:ga_distribution_joint}
    \end{subfigure}
    \caption{Distribution plots for returns and purchase decisions across episodes}
    \label{fig:ga_distribution_retpurjoint}
\end{figure}

In this section we present an overview of the distribution of returns and actions for the solved episodes. The objective is to show the performance of the agent across each solved episode and gauge its overall performance across 4,245 episodes. Another objective is to see any evident biases in action behavior or returns that we would not want our agent to have.
Figure \ref{fig:ga_distribution_returns} shows the distribution of the agent's return over daily average prices for the episode, this is calculated as:

\begin{equation}
    \left(1-\dfrac{\left(\dfrac{\vec{p} \cdot \vec{a}}{N_{a=2}}\right)}{\overline{p}}\right) * 100
\end{equation}

We can see from Figure \ref{fig:ga_distribution_returns} that about 2,100 episodes (~50\%) of the episodes have a return between 1-2\% over the daily average investment plan and about 3,800 episodes (~75\%) have a return of more than 1\% over the daily average. About 1,200 (25\%) of episodes have returns between 0-1\% over the daily average. We believe this behavior of the agent is highly desirable, since it generates a positive return of more than 1\% on a majority of occasions.

Figure \ref{fig:ga_distribution_purchasedecision} shows the purchasing decisions made by the agent over different episodes, where the x-axis represents the number of occasions on which the agent decided to buy (twice) within the 30 trading days. We can see from the graph that the agents purchase decisions are normally distributed, with a mean around 15 days. This is important, since it shows that our loss function is able to optimize to an average purchase decision of 15 of the 30 days. This means in the long run, on average, our agent would have purchased twice 15 times over a 30 trading day window and hence will have invested, in the long-run, the same amount as a daily investor. Figure 2 also shows another nice property of the distribution, most purchase decisions are between 11 - 19 trading days (~90\%), which shows that the agent is not over or under-purchasing during most episodes to maximize returns. This also shows that, it is very unlikely that an agent will make 30 consecutive buy decisions across multiple episodes, even though it is theoretically possible. We can control this behaviour by further tuning our loss function to truncate consecutive purchase decisions, though it would compromise returns. We leave this topic for future discussion and consideration.

Finally, Figure \ref{fig:ga_distribution_joint} shows the joint distribution of purchase occasions and returns to determine if higher returns are biased towards fewer purchase decisions. While independently, the distributions make sense, there might be certain unwanted behavior of the agent that may be uncovered in joint distributions. We can see from the joint distribution plot that abnormal returns of more than 5\% are not clustered towards low purchase decisions (less than 10 purchases), but are more or less evenly distributed. This exhibits another desirable behavior of our agent. 

We can see from above plots that we are able to solve for each episode independently with our loss function, yet able to produce results that, on average, would be able to maintain a portfolio size equal to the daily purchase behavior, while still attaining a return that is higher than the daily average strategy. Our next step is to determine if we can train a neural network model (a policy function) that can learn the behavior of this agent across episodes.

\section{Additional Details on GADLE Neural Network Performance (DL Part of GADLE)}
\label{appendix:additional NN details}

\begin{figure}[htbp]
    \centering
    \begin{subfigure}[t]{0.47\textwidth}
        \centering
        \includegraphics[width=\textwidth]{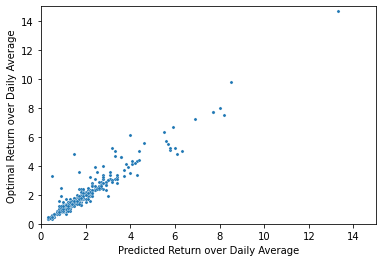}
        \caption{Predicted versus optimal return}
        \label{fig:pvoreturn}
    \end{subfigure}
    \begin{subfigure}[t]{0.47\textwidth}
        \centering
        \includegraphics[width=\textwidth]{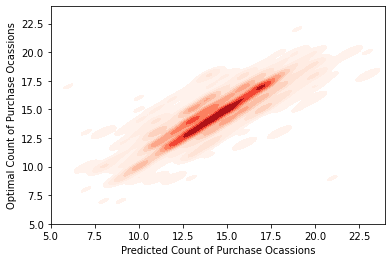}
        \caption{Predicted versus optimal actions}
        \label{fig:pvoaction}
    \end{subfigure}
    \caption{Performance plots for Deep Learning model training}
    \label{fig:pvoactionreturn}
\end{figure}

We present detailed results from our decoupled models performance in this section. Model accuracies presented in Section 2.6 show that our neural network is able to learn an overall policy function that closely resembles the actions taken by the GA to solve each episode. This fact can be visually seen in Figures \ref{fig:pvoreturn} and \ref{fig:pvoaction}.

Figure \ref{fig:pvoreturn} shows that the optimal return over the daily average obtained through the GA is very closely aligned with the returns obtained from a neural trained network model on the holdout test set of 500 episodes.

Similarly Figure \ref{fig:pvoaction} shows that the density of predicted versus optimal purchase decisions are very closely aligned, as expected, given our accuracy.

\section{DQN and Actor Critic RL Agent Training and Results}
\label{appendix:DQN and actor}

\subsection{DQN Agent Training}

DQN or Deep Q Network is one of the most basic deep RL algorithms that are in use today. This algorithm started the Deep Reinforcement Learning era in 2015. The major driving factor behind using this algorithm is the simplicity of the implementation and the widespread use of Q Learning, which is the most common non-deep RL algorithms. This algorithm involves using deep neural networks as a function to predict Q values given a state and action. This Q value represents the effectiveness of the action taken in that state. This information is transferred to the network by performing gradient descent using loss obtained from the discounted reward function. The full algorithm used to train the DQN agent is also mentioned below: 

\begin{algorithm}[htbp]
\SetAlgoLined
\caption{Deep Q-Learning}
$\mathbf{Q_\theta}$, a deep network parametrized by $\theta$ that accepts state, action pairs and returns a value estimate, initialized randomly\\
N, the number of episodes\\
\For{$i\gets0$ \KwTo $N$}{
    \For{$t\gets0$ \KwTo $30$}{
    With probability $\epsilon$, take random action $a_t \in \{0, 2\}$\\
    Otherwise $a_t = \arg\max_{a}\mathbf{Q_\theta}(s_t, a_t)$\\
    Take action $a_t$, observe reward $r_t$ and new state $s_{t+1}$\\
    Store examples in replay buffer\\
    Perform gradient descent on $\mathbf{Q}_\theta$
    }    
}
\label{algo:DQN Algorithm}
\end{algorithm}

\begin{table}[ht]
\centering
\caption{Parameters for DQN training}
\label{tab:dqn-parameters}
\begin{small}
\begin{sc}
\begin{tabular}{llc}
\toprule
Parameter & Definition & Value \\
\midrule
{Experience Replay} & Buffer size used & 7500 \\
& Unusual sampling factor & 0.9 \\
& Batch size sampled from the experience replay & 32 \\
\cmidrule{2-3}
{Learning Rate} & Initial Learning Rate & 0.001 \\
& Exponential Decay Steps & 1000 \\
& Exponential Decay Rate & 0.99 \\
& Exponential Decay Staircase & True \\
\cmidrule{2-3}
{Epsilon} & Initial Exploration Rate & 1.0 \\
& Minimum Exploration Rate & 0.01 \\
& Exploration Decay Rate (per episode) & 0.999\\
\midrule
Total episodes & Number of episodes (capped at 24 hr runtime) & 1440 \\
Batch Size & Size of batch for Q network training & 32 \\
Discount Factor & Discount factor for future rewards & 0.95 \\
Target Sync & Episodes to sync target and Q network weights & 2 \\
\midrule
\end{tabular}
\end{sc}
\end{small}
\end{table}

\begin{table}[htbp]
\centering
\caption{DQN Agent return and purchase decisions over daily SIP}
\label{tab:dqnresults}
\begin{small}
\begin{sc}
\begin{tabular}{lccccccc}
\toprule
{2020 Periods} & \multicolumn{3}{c}{Average Price} & & \multicolumn{3}{c}{No. of purchases} \\ \cline{2-4} \cline{6-8}
& Agent & Daily & RoD & & Agent & Daily & PCoD \\
\midrule
Jan 2nd - Feb 13th  & 163.43 & 161.98 & -0.89 \% & & 30 & 30 &   0 \\
Feb 14th - Mar 27th & 125.49 & 139.38 &  9.96 \% & & 34 & 30 &   4 \\
Mar 30th - May 11th & 136.27 & 135.13 & -0.85 \% & & 36 & 30 &   6 \\
May 12th - Jun 23rd & 146.05 & 149.88 &  2.56 \% & & 18 & 30 & -12 \\
Jun 24th - Aug 5th  & 160.67 & 157.93 & -1.74 \% & & 30 & 30 &   0 \\
Aug 6th - Sep 17th  & 170.18 & 169.10 & -0.64 \% & & 40 & 30 &  10 \\
Sep 18th - Oct 29th & 170.98 & 169.61 & -0.81 \% & &  2 & 30 & -28 \\
Oct 30th - Dec 11th & 182.50 & 180.83 & -0.92 \% & & 50 & 30 &  20 \\
\midrule
Overall             & 157.49 & 157.98 &  0.31 \% & & 240 & 240 & 0 \\
\bottomrule
\end{tabular}
\end{sc}
\end{small}
\end{table}

We implemented this algorithm with the same neural network architecture as used in the GADLE method to ensure same number of trainable parameters in each. We also modified the experience replay to give more weightage to non-zero reward samples while sampling data for training. A unusual sampling factor of 0 would cause extreme weightage shift and would only return non-zero reward samples, while an unusual sampling factor value of 1 would correspond to uniform sampling. After performing some tests, we set the value of this parameter to 0.9. Apart from this, we are also using a learning rate scheduler and an epsilon decay scheme. The details of these and the rest of the parameters used for training are mentioned in Table \ref{tab:dqn-parameters}.

The training for DQN was observed to be very slow as compared to GADLE method. It took roughly 1 minute to train for a single episode on an NVIDIA Tesla P100. Hence, for computational reasons, we decided to keep the upper limit of training time to be 24 hours. This capping corresponded to a total of 1440 episodes for DQN. Another thing that we observed for DQN was it was not very consistent in its runs and the run heavily depended on the hyperparameters. Even a small change could lead to the run failing , which we defined as the run having either a buy ratio of 0 (not purchasing anything) or 1 (purchasing everything, effectively equal to SIP). We would discuss this in much more detail in further sections.

After analysing the results for the completed runs, we found that even after implementing the exact same reward function structure as the GADLE and after doing some hyperparameter tuning, the results were not satisfactory. The performance of the DQN model on the 2020 data can be seen in Table \ref{tab:dqnresults}. It can be seen that the Return over Daily is just 0.31\% as compared to the 1.26\% observed in GADLE.

\begin{figure}[htbp]
    \centering
    \begin{subfigure}[t]{0.45\textwidth}
        \centering
        \includegraphics[width=\textwidth]{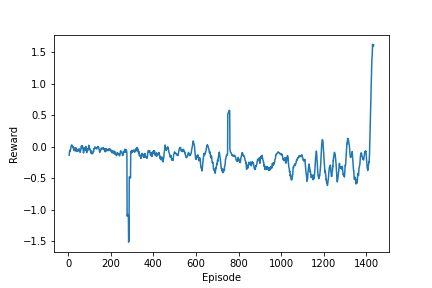}
        \caption{Running reward over episodes}
        \label{fig:dqn_training_plots_rr}
    \end{subfigure}
    \begin{subfigure}[t]{0.45\textwidth}
        \centering
        \includegraphics[width=\textwidth]{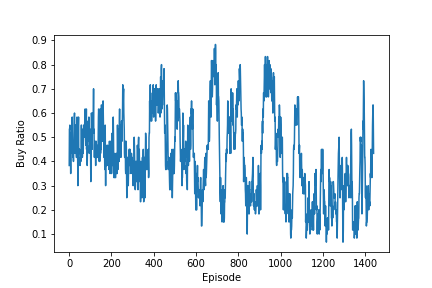}
        \caption{Buy ratio over episodes}
        \label{fig:dqn_training_plots_br}
    \end{subfigure}
    \caption{Training plots for DQN agent}
    \label{fig:dqn_training_plots}
\end{figure}

The training plots for DQN training can be seen in Figure \ref{fig:dqn_training_plots}. They also show a very similar picture. It can be seen in Figure \ref{fig:dqn_training_plots_rr} that the reward is not going up, and in general, have a slightly downward trend. It can also be noticed that it does seem to rise sharply towards the end of the episode, but it has been noticed that such peaks are short lived (as can bee seen around Episodes 250 and 800). The buy ratio plot seen in Figure \ref{fig:dqn_training_plots_br} also doesn't show any convergence to the optimal buy value which is 0.5 in our case.

\subsection{Actor-Critic Agent Training}

Actor-Critic Agents are one of the most common agent based deep RL techniques in use today. The asynchronous version of this algorithm (A3C) gives state of the art results for most benchmark problems in RL. This algorithm involves 2 different agents, actor and critic. The role of the actor is to return probabilities for each action in the action space given the state based on the 'goodness' of the action. On the other hand, the role of the critic is to judge the action taken by the actor, hence encouraging the actor to select the actions with better and better rewards.

\begin{algorithm}[htb]
\SetAlgoLined
\caption{{Modified Advantage Actor-Critic (A2C)}}
$\mathbf{A_\theta}$, a deep actor network parametrized by $\theta$ that accepts state and returns action probabilities for each possible action\\
$\mathbf{C_\theta}$, a deep critic network parametrized by $\theta$ that  returns a critic estimate\\
(Both $\mathbf{A_\theta}$ and $\mathbf{C_\theta}$ share the same backbone and are initialised randomly)\\
$\mathbf{\gamma} = 0.97$, discount factor\\
$\mathbf{r_{R}} = 0$, running reward\\
$\mathbf{r_{Rmax}} = -\infty$, maximum running reward\\
$\mathbf{\epsilon} = 1.0$, exploration rate\\
$\mathbf{\epsilon_{min}} = 0.01$, minimum exploration rate\\
$\mathbf{\epsilon_{decay}} = 0.999$, exploration rate decay factor\\
\For{forever until not solved}{
    Sample a random episode and initialise $s_t = s_0$\\
    \For{$t\gets0$ \KwTo $30$}{
        Get action probabilities vector $\vec{a}_{t} = \mathbf{A_\theta}(s_t)$\\
        Get critic value $c_t = \mathbf{C_\theta}(s_t)$\\
        With probability $\epsilon$, take action $a_t \in \{0, 2\}$ based on $\vec{a}_{t}$\\
        Otherwise $a_t = \arg\max_{a}\vec{a}_{t}$\\
        Take action $a_t$, observe reward $r_t$ and new state $s_{t+1}$\\
        Store $\log a_t$, $c_t$ and $r_t$ in history $H(a,c,r)$\\
    }
    If $\mathbf{\epsilon} > \mathbf{\epsilon_{min}}$ then $\mathbf{\epsilon} = \mathbf{\epsilon} * \mathbf{\epsilon_{decay}}$\\
    $\mathbf{r_{R}} = 0.05 * \mathbf{r_{episode}} + 0.95 * \mathbf{r_{R}}$\\
    Replace $H(r)$ with normalised discounted rewards using $\mathbf{\gamma}$\\
    \For{$log\_prob, value, return \gets H(a,c,r)$}{
        Advantage $adv = return - value$\\
        Actor Loss $L_{A} = -log\_prob * adv$\\
        Critic Loss $L_{C} = L_{H}(value, return)$ where $L_{H}$ is the Huber loss function\\
    }
    Perform gradient descent on $\mathbf{A}_\theta$, $\mathbf{C}_\theta$\\
    Clear history $H(a,c,r)$
}
\label{algo:Modified Actor-Critic}
\end{algorithm}

Here, we have used a variation of the Actor-critic algorithm where the actor and the critic share the same backbone. This was done to ensure that the neural network structure remains the same across our different methods. Also, we tried out vanilla Actor-critic and on inspection we found that it was performing well below par due to some exploration issues. Hence, we modified it a little bit to have a controlled exploration rate, guided by a decaying epsilon, much like the epsilon greedy technique. The full algorithm used to train the Actor-critic agent is also shown for better clarity.

Apart from this, we have also used an exponential learning rate decay function for the actor-critic network training. Also, the epsilon-decay and learning rate parameters are kept exactly the same as those used while training the DQN agent. The details of these and the rest of the parameters can be seen in Table \ref{tab:a2c-parameters}.

\begin{table}[htbp]
\centering
\caption{Parameters for Actor-Critic training}
\label{tab:a2c-parameters}
\begin{small}
\begin{sc}
\begin{tabular}{llc}
\toprule
Parameter & Definition & Value \\
\midrule
{Learning Rate} & Initial Learning Rate & 0.001 \\
& Exponential Decay Steps & 1000 \\
& Exponential Decay Rate & 0.99 \\
& Exponential Decay Staircase & True \\
\cmidrule{2-3}
{Epsilon} & Initial Exploration Rate & 1.0 \\
& Minimum Exploration Rate & 0.01 \\
& Exploration Decay Rate (per episode) & 0.999\\
\midrule
Discount Factor & Discount factor for future rewards & 0.97 \\
\bottomrule
\end{tabular}
\end{sc}
\end{small}
\end{table}

The training for Actor-critic was much faster than DQN training. However, since RL algorithms need some time to explore the action space and slowly learn the optimum actions for each state space, it was expected that the algorithm would take more time than the GADLE method to get comparable results. However, to keep the comparison just between the different algorithms, we decided to cap the training time for this too at 24 hours, like the DQN agent. However, since Actor-critic was much faster, this limit corresponded to roughly 288,000 episodes.

After analysing the results for the completed runs, we observed that the performance of Actor-critic was slightly better than our proposed GADLE. The evaluation of the agent on the 2020 data can be seen in Table \ref{tab:a2cresults}. Actor-critic agent was able to achieve an RoD value of 1.38\% compared to the 1.26\% of GADLE. The PCoD was also reasonable coming down to an average of -1.25 purchase difference per 30 purchases made by daily. This would average out to zero in the long run, as explained earlier.

\begin{table}[htbp]
\centering
\caption{Actor-Critic Agent return and purchase decisions over daily SIP}
\label{tab:a2cresults}
\begin{small}
\begin{sc}
\begin{tabular}{lccccccc}
\toprule
{2020 Periods} & \multicolumn{3}{c}{Average Price} & & \multicolumn{3}{c}{No. of purchases} \\ \cline{2-4} \cline{6-8}
& Agent & Daily & RoD & & Agent & Daily & PCoD \\
\midrule
Jan 2nd - Feb 13th  & 161.80 & 161.98 &  0.11 \% & & 26 & 30 & -4 \\
Feb 14th - Mar 27th & 134.27 & 139.38 &  3.66 \% & & 46 & 30 & 16 \\
Mar 30th - May 11th & 136.01 & 135.13 & -0.65 \% & & 26 & 30 & -4 \\
May 12th - Jun 23rd & 150.14 & 149.88 & -0.17 \% & & 24 & 30 & -6 \\
Jun 24th - Aug 5th  & 158.67 & 157.93 & -0.47 \% & & 28 & 30 & -2 \\
Aug 6th - Sep 17th  & 168.86 & 169.10 &  0.14 \% & & 28 & 30 & -2 \\
Sep 18th - Oct 29th & 171.02 & 169.61 & -0.83 \% & & 28 & 30 & -2 \\
Oct 30th - Dec 11th & 181.30 & 180.83 & -0.26 \% & & 24 & 30 & -6 \\
\midrule
Overall             & 155.80 & 157.98 &  1.38 \% & & 230 & 240 & -10 \\
\bottomrule
\end{tabular}
\end{sc}
\end{small}
\end{table}

\begin{figure}[!ht]
    \centering
    \begin{subfigure}[t]{0.45\textwidth}
        \centering
        \includegraphics[width=\textwidth]{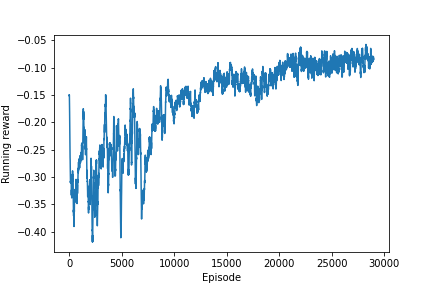}
        \caption{Running reward over episodes}
        \label{fig:a2c_training_plots_rr}
    \end{subfigure}
    \begin{subfigure}[t]{0.45\textwidth}
        \centering
        \includegraphics[width=\textwidth]{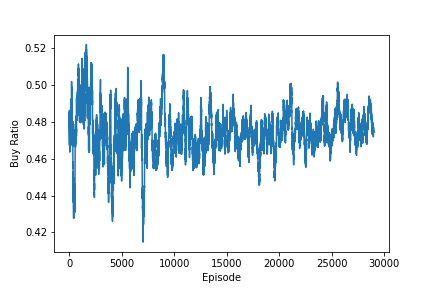}
        \caption{Buy ratio over episodes}
        \label{fig:a2c_training_plots_br}
    \end{subfigure}
    \caption{Training plots for Actor-critic agent}
    \label{fig:a2c_training_plots}
\end{figure}

The training plots for the Actor-critic agent can be seen in Figure \ref{fig:a2c_training_plots}. They are very close to the ideal plots we expected from an RL method. In Figure \ref{fig:a2c_training_plots_rr} we can see that the reward is going steadily up with episodes. It can be seen that there were some fluctuations in the beginning which can be attributed to the high exploration rate enforced by our modified algorithm. Also, it can also be seen in Figure \ref{fig:a2c_training_plots_br} that the buy ratio is converging slowly close to the optimum value of 0.5.

\section{Sensitivity runs for all algorithms}
\label{appendix:sensitivity runs}

The sensitivity results for GADLE methods are present in Table \ref{tab:ga_sensitivity}. Several parameters pertaining to GA have been either increased or decreased by 20 \% from the baseline values as present in Table \ref{tab:ga_parameters}. All of these have been re-trained and evaluated on the 2020 data. It can be seen from the table that the performance of all these variations are very comparable to that of our original run, the baseline. One thing to note is that no run was stuck or gave outrageous results to be classified as failed. To quantify this, we also calculated the absolute average deviation of all these different variations with baseline and presented in the table. The Agent Average price deviation comes out to be 0.51 and that of RoD (return over daily) and PCoD (purchase count over daily averaged over a month) comes out to be 0.33 \% and 1.42 respectively. These are very less and are easily understandable. Hence, we can say that our proposed GADLE method is quite stable with respect to its parameters.

The sensitivity results for Actor-critic method are present in Table \ref{tab:a2c_sensitivity}. It can be seen that 2 runs failed wherein we increased the initial learning rate and increased the learning rate decay rate. Also, other runs which did successfully run, the variation in performance is huge as compared to GADLE. This can also be seen in the values of the average absolute deviations. The RoD and PCoD variation is 1.75 \% and 4.61 respectively, which compared to the corresponding values for GADLE method are extremely high. This shows that Actor-critic method training is very sensitive to change in its parameters.

\begin{table*}[ht]
\centering
\caption{GADLE Agent performance variation with change in parameters}
\label{tab:ga_sensitivity}
\begin{small}
\begin{sc}
\begin{tabular}{lcccr}
\toprule
Parameter Change & Agent Avg. & Daily Avg. & RoD & PCoD \\
\midrule
Baseline                  & 155.99 & 157.98 & 1.26 \% & -1.25 \\
\midrule
population size dec       & 156.16 & 157.98 & 1.15 \% & -0.5 \\
population size inc       & 156.89 & 157.98 & 0.69 \% & -2.75 \\
crossover probability dec & 156.10 & 157.98 & 1.19 \% & -1.75 \\
crossover probability inc & 156.10 & 157.98 & 1.19 \% & -3.0 \\
mutation probability dec  & 155.10 & 157.98 & 1.82 \% & -2.5 \\
mutation probability inc  & 157.09 & 157.98 & 0.56 \% & -5.0 \\
elite ratio dec           & 156.25 & 157.98 & 1.10 \% & -1.0 \\
parents portion inc       & 156.51 & 157.98 & 0.93 \% & -0.75 \\
crossover type two point  & 156.24 & 157.98 & 1.10 \% & 0.75 \\
crossover type one point  & 156.81 & 157.98 & 0.74 \% & -3.25 \\
\midrule
Avg. Absolute Deviation   & 0.51   & 0.0    & 0.33 \% & 1.42 \\
\bottomrule
\end{tabular}
\end{sc}
\end{small}
\end{table*}

\begin{table*}[ht]
\centering
\caption{Actor-critic Agent performance variation with change in parameters}
\label{tab:a2c_sensitivity}
\begin{small}
\begin{sc}
\begin{tabular}{lcccr}
\toprule
Parameter Change & Agent Avg. & Daily Avg. & RoD & PCoD \\
\midrule
Baseline                  & 155.80 & 157.98 & 1.38 \% & -1.25 \\
\midrule
initial $\epsilon$ dec & 158.95 & 157.98 & -0.61 \% & -0.75 \\
minimum $\epsilon$ dec & 159.14 & 157.98 & -0.73 \% & 2.75 \\
minimum $\epsilon$ inc & 160.72 & 157.98 & -1.73 \% & -12.75 \\
$\epsilon$ decay rate dec & 160.51 & 157.98 & -1.60 \% & -8.75 \\
$\epsilon$ decay rate inc & 157.76 & 157.98 & 0.14 \% & 7.25 \\
initial lr rate dec & 155.13 & 157.98 & 1.80 \% & -1.75 \\
initial lr rate inc & FAIL & FAIL & FAIL & FAIL \\
lr rate steps dec & 158.58 & 157.98 & -0.38 \% & -3.0 \\
lr rate steps inc & 156.33 & 157.98 & 1.04 \% & -6.75 \\
lr decay rate dec & 158.68 & 157.98 & -0.44 \% & 0.5 \\
lr decay rate inc & FAIL & FAIL & FAIL & FAIL \\
\midrule
Avg. Absolute Deviation   & 2.77   & 0.0    & 1.75 \% & 4.61 \\
\bottomrule
\end{tabular}
\end{sc}
\end{small}
\end{table*}

We haven't included the sensitivity results for the DQN method as it was so unstable, that even a slight change in its parameters lead to the run crashing and the model obtained either purchased nothing or everything.

To summarise, we would say that our proposed GADLE method is much more stable with respect to its parameters than Actor-critic method and also, the DQN agent is extremely volatile. Hence, conversely, finding optimal parameters is much easier in GADLE method than the other 2 RL-based methods.

\section{Consistency runs for all algorithms}
\label{appendix:consistency runs}

As shown in section \ref{sec:consistency}, GADLE method performs in a manner more reliable when initiated with a random seed as compared to DQN and Actor-Critic. In this section we show further details and plots for the same.

\begin{figure}[t]
    \centering
    \begin{subfigure}[t]{0.47\textwidth}
        \centering
        \includegraphics[width=\textwidth]{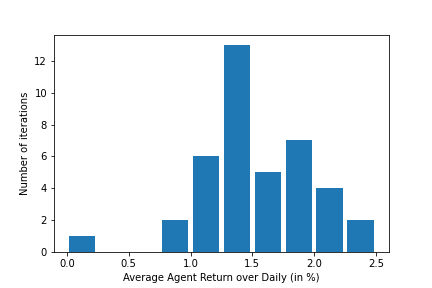}
        \caption{RoD Distribution}
        \label{fig:ga_consistency_plots_rod}
    \end{subfigure}
    \begin{subfigure}[t]{0.47\textwidth}
        \centering
        \includegraphics[width=\textwidth]{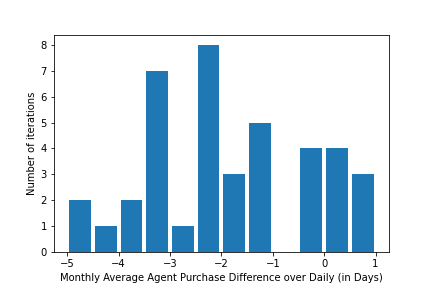}
        \caption{PCoD Distribution}
        \label{fig:ga_consistency_plots_pcod}
    \end{subfigure}
    \caption{GADLE agent performance variation for consistency runs}
    \label{fig:ga_consistency_plots}
\end{figure}

The GADLE method was found to be very robust in terms of returns, with all 40 iterations performing better than the daily agent and more than 90\% of the iterations yielding returns greater than 1\% over daily agent as seen in Figure \ref{fig:ga_consistency_plots_rod}. Also, the agent is purchasing almost equal number of times as the daily SIP as can be seen in Figure \ref{fig:ga_consistency_plots_pcod}. However, there is a slight under-purchasing trend that can be observed in some of the cases, but it can be observed that more than 50 \% of the iterations are within the $\pm$ 2 PCoD region. Hence, it can be said that the level of under-purchase is very minute and will average out to 0 over long term, as demonstrated earlier.

\begin{figure}[t]
    \centering
    \begin{subfigure}[t]{0.47\textwidth}
        \centering
        \includegraphics[width=\textwidth]{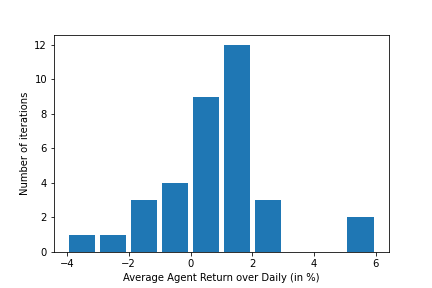}
        \caption{RoD Distribution}
        \label{fig:a2c_consistency_plots_rod}
    \end{subfigure}
    \begin{subfigure}[t]{0.47\textwidth}
        \centering
        \includegraphics[width=\textwidth]{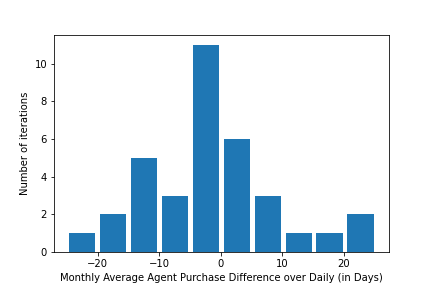}
        \caption{PCoD Distribution}
        \label{fig:a2c_consistency_plots_pcod}
    \end{subfigure}
    \caption{Actor-critic agent performance variation for consistency runs}
    \label{fig:a2c_consistency_plots}
\end{figure}

The variation of performance for the Actor-critic agent can be seen in Figure \ref{fig:a2c_consistency_plots}. It should be taken into account that 5 of the total 40 runs were marked unsuccessful as those runs had below par reward progressions as compared to the others. Also, the reward was either not increasing at all, or was too erratic to be considered a stable run. The Actor-critic agent was found to be significantly worse than our proposed GADLE method in robustness. Roughly 25 \% of the runs were found to be worse than the daily agent and only about 50 \% of the runs were found to yield RoD value greater than 1 \% as can be seen in Figure \ref{fig:a2c_consistency_plots_rod}. Also, it was found that roughly 50 \% of the runs lied within the $\pm$ 5 PCoD region which is acceptable at best.

\begin{figure}[p]
    \centering
    \begin{subfigure}[t]{0.4\textwidth}
        \centering
        \includegraphics[width=\textwidth]{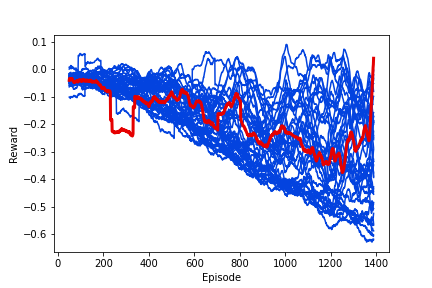}
        \caption{Reward over episodes for all runs}
        \label{fig:dqn_consistency_overepisodes_all_rr}
    \end{subfigure}
    \begin{subfigure}[t]{0.4\textwidth}
        \centering
        \includegraphics[width=\textwidth]{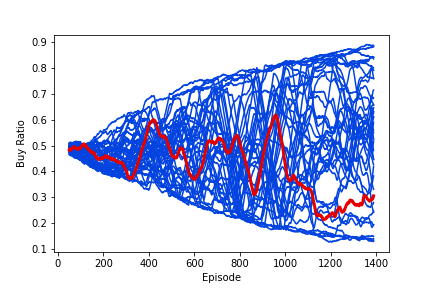}
        \caption{Buy ratio over episodes for all runs}
        \label{fig:dqn_consistency_overepisodes_all_br}
    \end{subfigure}
    \vskip -0.2in
    \begin{subfigure}[t]{0.4\textwidth}
        \centering
        \includegraphics[width=\textwidth]{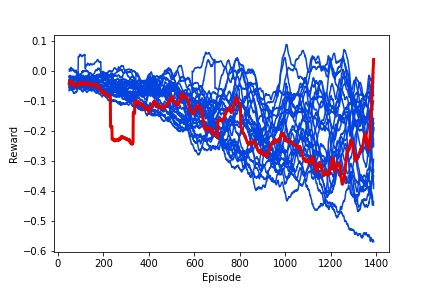}
        \caption{Reward over episodes for successful runs}
        \label{fig:dqn_consistency_overepisodes_pruned_rr}
    \end{subfigure}
    \begin{subfigure}[t]{0.4\textwidth}
        \centering
        \includegraphics[width=\textwidth]{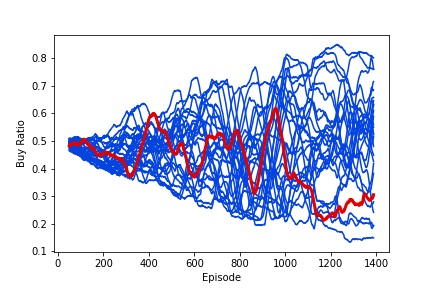}
        \caption{Buy ratio over episodes for successful runs}
        \label{fig:dqn_consistency_overepisodes_pruned_br}
    \end{subfigure}
    \caption{DQN agent performance variation for consistency runs. Plots (a) and (b) show the plots for all 40 consistency runs while (c) and (d) prune 14 failed runs and only show successful runs. The consistency runs (in blue) are overlaid with the original run (in red).}
    \label{fig:dqn_consistency_overepisodes}
\end{figure}

\begin{figure}[p]
    \centering
    \begin{subfigure}[t]{0.4\textwidth}
        \centering
        \includegraphics[width=\textwidth]{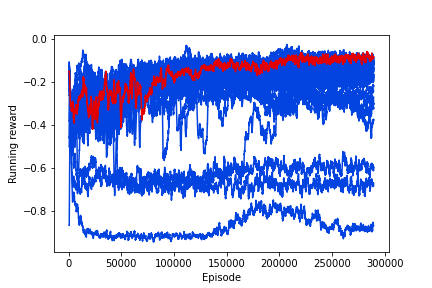}
        \caption{Running reward over episodes for all runs}
        \label{fig:a2c_consistency_overepisodes_all_rr}
    \end{subfigure}
    \begin{subfigure}[t]{0.4\textwidth}
        \centering
        \includegraphics[width=\textwidth]{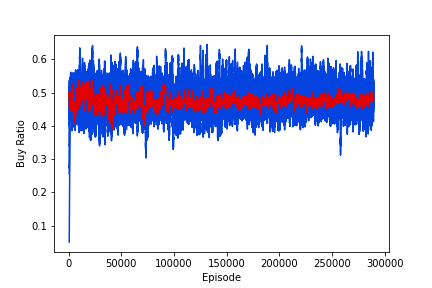}
        \caption{Buy ratio over episodes for all runs}
        \label{fig:a2c_consistency_overepisodes_all_br}
    \end{subfigure}
    \vskip -0.2in
    \begin{subfigure}[t]{0.4\textwidth}
        \centering
        \includegraphics[width=\textwidth]{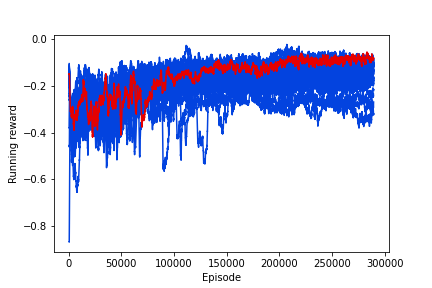}
        \caption{Running reward over episodes for successful runs}
        \label{fig:a2c_consistency_overepisodes_pruned_rr}
    \end{subfigure}
    \begin{subfigure}[t]{0.4\textwidth}
        \centering
        \includegraphics[width=\textwidth]{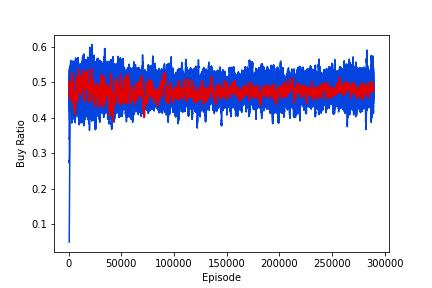}
        \caption{Buy ratio over episodes for successful runs}
        \label{fig:a2c_consistency_overepisodes_pruned_br}
    \end{subfigure}
    \caption{Actor-critic agent performance variation for consistency runs. Plots (a) and (b) show the plots for all 40 consistency runs while (c) and (d) prune 5 failed runs and only show successful runs. The consistency runs (in blue) are overlaid with the original run (in red).}
    \label{fig:a2c_consistency_overepisodes}
\end{figure}

We further show the consistency of results by plotting the training plots for both DQN and Actor-critic. It can be seen from the reward in figure \ref{fig:dqn_consistency_overepisodes} and \ref{fig:a2c_consistency_overepisodes}, plots (a) and (c) that the general trend for all the runs is increasing rewards which is expected and good, however, for both DQN and Actor-Critic, there are certain runs with negative rewards - which we categorize as failed runs. Further, it can be observed that the low-hanging reward runs from the plots are removed in the pruned version (figure \ref{fig:dqn_consistency_overepisodes} and \ref{fig:a2c_consistency_overepisodes}, plot (c)). Finally, it can be seen from the pruned version of the plot a conical convergence general theme emerging from the runs depicting convergence to a near 0.5 value (figure \ref{fig:dqn_consistency_overepisodes} and \ref{fig:a2c_consistency_overepisodes}, plot (d)). 

\vskip 1in

\section{Production Architecture for GADLE testing}
\begin{figure}[htb]
\centering
  \centerline{\includegraphics[width=0.7\linewidth]{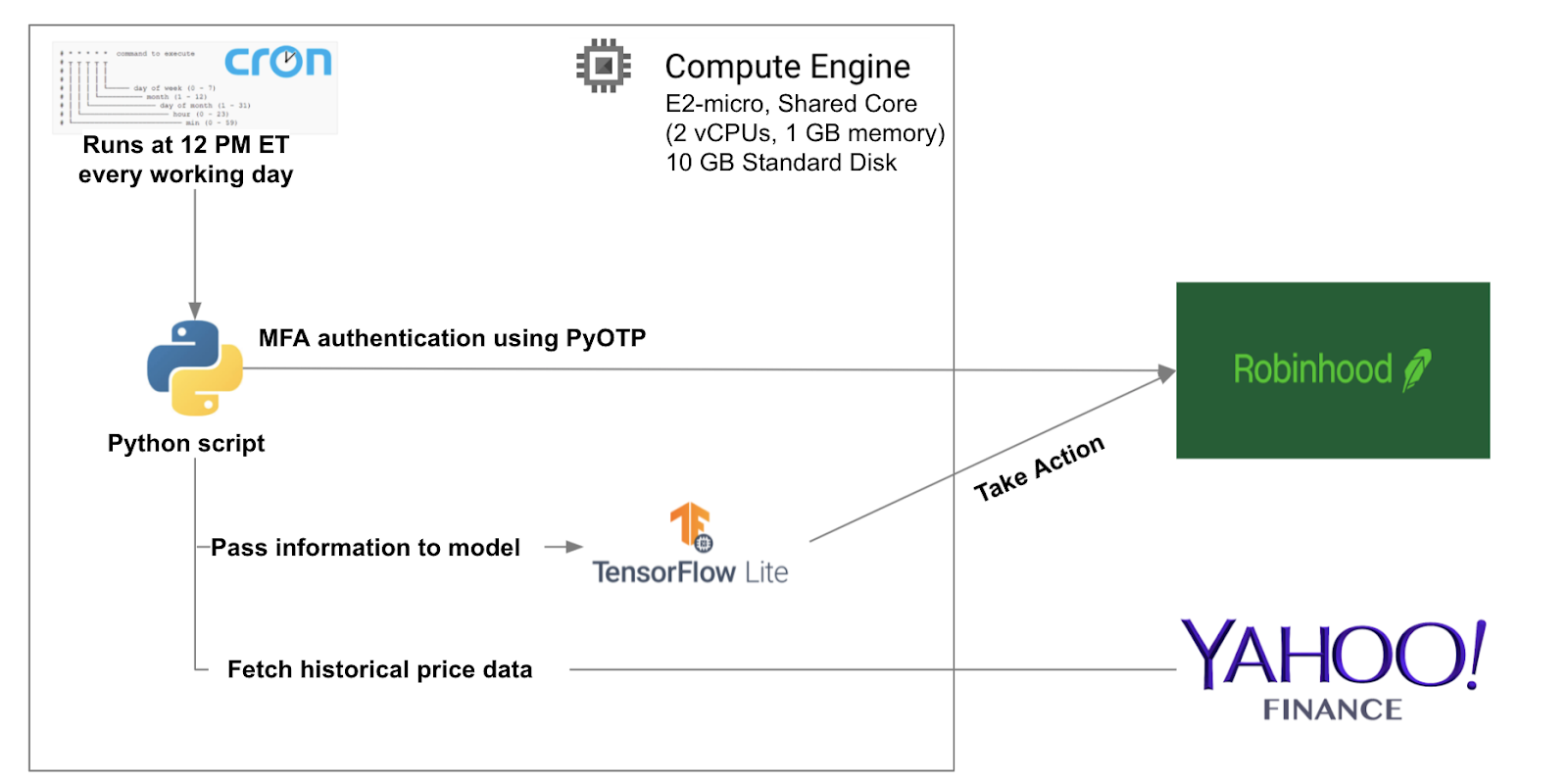}}
  \caption{Architecture for executing our algorithm in production environment}
  \label{fig:architecture for agent}
  
\end{figure}

We present in this section the architecture diagram used to execute our agent decisions in production environment in figure \ref{fig:architecture for agent}.
\label{appendix:production_architecture}

\end{document}